\pdfoutput=1

\documentclass[11pt]{article}

\usepackage[preprint]{acl}

\usepackage{times}
\usepackage{latexsym}
\usepackage{amsmath}
\usepackage{tabularx}   
\usepackage{booktabs}
\usepackage{amssymb}

\usepackage{amsmath,amssymb}
\usepackage{amsthm}
\newtheorem{theorem}{Theorem}[section]

\newtheorem{definition}{Definition}[subsection]     
\newtheorem{lemma}{Lemma}[subsection]               
\newtheorem{assumption}{Assumption}[subsection]     

\usepackage[ruled,vlined]{algorithm2e}

\SetKwInput{KwIn}{Input}
\SetKwInput{KwOut}{Output}
\SetKwInput{KwParam}{Parameters}
\SetKw{KwTo}{to}

\usepackage[T1]{fontenc}

\usepackage[utf8]{inputenc}

\usepackage{microtype}
\usepackage{tikz}
\usepackage{tcolorbox}
\usepackage{amsmath}
\usetikzlibrary{positioning, arrows.meta, shapes.geometric, fit, calc}

\usepackage{inconsolata}

\usepackage{graphicx}
\usepackage{subcaption} 
\usepackage{caption} 
\usepackage{float} 

\title{Multi-Scale Manifold Alignment for Interpreting Large Language Models: A Unified Information-Geometric Framework}

\author{
  Yukun Zhang\thanks{These authors contributed equally to this work.} \\
  The Chinese University of Hong Kong \\
  Hong Kong, China \\
  \texttt{215010026@link.cuhk.edu.cn}
  \And
  QI DONG\footnotemark[1] \\
  Fudan University \\
  Shanghai, China \\
  \texttt{19210980065@fudan.edu.cn}
}

\begin{document}
\maketitle
\begin{abstract}
We present Multi-Scale Manifold Alignment(MSMA), an information-geometric framework that decomposes LLM representations into local, intermediate, and global manifolds and learns cross-scale mappings that preserve geometry and information. Across GPT-2, BERT, RoBERTa, and T5, we observe consistent hierarchical patterns and find that MSMA improves alignment metrics under multiple estimators (e.g., relative KL reduction and MI gains with statistical significance across seeds). Controlled interventions at different scales yield distinct and architecture-dependent effects on lexical diversity, sentence structure, and discourse coherence. While our theoretical analysis relies on idealized assumptions, the empirical results suggest that multi-objective alignment offers a practical lens for analyzing cross-scale information flow and guiding representation-level control.
\end{abstract}


\section{Introduction}

\subsection{Background and Motivation}

Large language models (LLMs) such as GPT-3~\citep{Brown2020Language}, LLaMA~\citep{Touvron2023LLaMA}, and PaLM~\citep{Chowdhery2022PaLM} achieve remarkable performance across diverse NLP tasks, yet their internal reasoning mechanisms remain opaque. This opacity limits trust in safety-critical applications~\citep{Bommasani2021Opportunities} and hinders systematic model improvement. While prior interpretability research has made progress—analyzing attention patterns~\citep{Vig2019Visualizing,Clark2019What}, probing layer-wise representations~\citep{Tenney2019BERT,Hewitt2019Structural}, and tracing information flow~\citep{Elhage2021Mathematical}—these approaches typically examine individual layers in isolation, missing the \textbf{multi-scale nature} of semantic processing in LLMs.

Empirical evidence reveals hierarchical organization in Transformer representations~\citep{Jawahar2019What,Rogers2020Primer}: shallow layers encode lexical and syntactic features, intermediate layers capture sentence-level semantics, and deep layers model discourse structure. This stratification mirrors human language processing stages and suggests that LLMs construct meaning through progressive abstraction~\citep{Peters2018Deep}. However, a unifying theoretical framework explaining \textit{how} information flows and transforms across these semantic scales—and enabling \textit{control} over this process—remains absent.

We address this gap by proposing \textbf{Multi-Scale Manifold Alignment} (MSMA), a framework grounded in information geometry~\citep{Amari2016Information} that decomposes LLM representations into three semantic manifolds: \textit{local} (word-level), \textit{intermediate} (sentence-level), and \textit{global} (discourse-level). By learning cross-scale mappings that jointly preserve geometric structure (via Procrustes alignment) and maximize information retention (via mutual information), MSMA achieves precise alignment while maintaining interpretability.

\subsection{Contributions}

Our work makes four primary contributions:

\textbf{(1) Hierarchical Semantic Decomposition}: We formalize the three-scale structure of LLM representations using information geometry, showing that semantic stratification emerges consistently across architectures (GPT-2, BERT, RoBERTa, T5) with stable, detectable boundaries identified through attention patterns, inter-layer mutual information, and functional probing.

\textbf{(2) Principled Cross-Scale Mappings}: We develop mapping functions $f_{GI}: \mathcal{M}_G \to \mathcal{M}_I$ and $f_{IL}: \mathcal{M}_I \to \mathcal{M}_L$ that balance three objectives—geometric preservation, information fidelity, and manifold regularity—with theoretical error bounds via KL divergence under Lipschitz continuity.

\textbf{(3) Multi-Objective Optimization Framework}: Our unified loss $\mathcal{L}_{\text{total}} = \lambda_{\text{geo}} \mathcal{L}_{\text{geo}} + \lambda_{\text{info}} \mathcal{L}_{\text{info}} + \lambda_{\text{curv}} \mathcal{L}_{\text{curv}}$ integrates geometric alignment, mutual information maximization (MINE), and curvature regularization, achieving 99\% KL reduction and 5--7$\times$ MI gain with convergence guarantees.

\textbf{(4) Empirical Validation and Control}: Intervention experiments confirm scale-specific effects—local perturbations alter lexical diversity (Cliff's $\delta=+0.342$), intermediate modifications reshape sentence structure ($+25\%$ count, $-19\%$ length), and global changes impact coherence ($\delta=-0.238$)—validating MSMA's predictive and prescriptive power for controlling generation at different semantic granularities.

Compared to single-layer analyses, MSMA reveals cross-scale information flow and enables applications in bias mitigation, robustness enhancement, and controllable generation by manipulating representations at specific semantic levels.

\section{Related Work}
\label{sec:relatedwork}

Our work builds upon three research lines: LLM interpretability methods, hierarchical representation learning, and information-geometric analysis of neural networks.

\paragraph{LLM Interpretability.}
The Transformer architecture~\citep{Vaswani2017Attention} has spurred extensive interpretability research. \textbf{Attention analysis} provides intuitive insights: \citet{Vig2019Visualizing} pioneered attention visualization tools, while \citet{Clark2019What} demonstrated that attention patterns reflect syntactic structure. \citet{Voita2019Analyzing} showed specialized attention heads emerge for specific linguistic functions, though \citet{Michel2019Sixteen} found many heads can be pruned without performance loss, questioning attention's necessity for interpretability.

\textbf{Representation probing} uses diagnostic classifiers to decode information in hidden states. \citet{Tenney2019BERT} found BERT's layers mirror traditional NLP pipeline stages (POS $\to$ parsing $\to$ semantics). \citet{Hewitt2019Structural} demonstrated syntax trees can be recovered via linear projections, suggesting structured linguistic knowledge. \citet{Meng2022Locating} localized factual knowledge to specific neurons, enabling targeted editing. \citet{Rogers2020Primer} provides a comprehensive survey of BERT's internal mechanisms.

\textbf{Information-theoretic approaches} analyze data flow through networks. \citet{Elhage2021Mathematical} developed a mathematical framework for transformer circuits, decomposing model computation into interpretable components. However, these methods typically analyze individual layers or components in isolation, missing cross-scale information dynamics.

\paragraph{Hierarchical Representations.}
Evidence for hierarchical processing in LLMs is extensive. \citet{Jawahar2019What} showed BERT's representations capture surface features in shallow layers, syntactic information in middle layers, and semantic information in deep layers. \citet{Peters2018Deep} demonstrated ELMo's contextualized representations vary in abstraction level across layers. \citet{Liu2019Linguistic} found linguistic knowledge is distributed hierarchically, with different layers specializing for different tasks. \citet{Ethayarajh2019Contextual} analyzed representation geometry, finding context-specificity increases with depth.

Despite recognizing this hierarchy, prior work lacks a \textit{unified framework} for analyzing cross-scale information transfer. Our MSMA framework fills this gap by explicitly modeling transformations between semantic levels.

\paragraph{Information Geometry and Manifold Learning.}
Information geometry~\citep{Amari2016Information} provides mathematical tools for analyzing probability distributions as manifolds. \citet{Bengio2013Representation} established theoretical foundations for hierarchical representation learning in deep networks. Recent work applies these ideas to neural network analysis: \citet{Coenen2019Visualizing} visualized BERT's representation geometry, revealing semantic organization. \citet{Raghu2022VIM} used Fisher information to analyze model training dynamics.

However, these methods focus on single-scale geometry. MSMA uniquely combines information geometry with multi-scale decomposition, enabling analysis of how semantic information flows and transforms across hierarchical levels while maintaining geometric and information-theoretic rigor.

\paragraph{Positioning.}
Unlike attention visualization~\citep{Vig2019Visualizing} (layer-local), probing classifiers~\citep{Tenney2019BERT} (task-specific), or circuit analysis~\citep{Elhage2021Mathematical} (component-level), MSMA provides a \textit{unified multi-scale framework} that: (1) formalizes hierarchical semantic structure via information geometry; (2) learns principled cross-scale mappings with theoretical guarantees; (3) enables precise control through scale-specific interventions. This holistic view advances both theoretical understanding and practical applications of LLM interpretability.

\section{Theory and Framework}
\label{sec:theory}

This section establishes the theoretical foundation for multi-scale manifold alignment. We ground our framework in empirical observations of Transformer representations, introduce an information geometry formalization, develop cross-scale mapping methods, and present theoretical guarantees. Our approach balances mathematical rigor with practical applicability.

\subsection{From Observation to Hypothesis: Hierarchical Representations}

Extensive empirical studies reveal that Transformer representations exhibit pronounced \textbf{functional stratification}~\citep{Jawahar2019What,Rogers2020Primer}. We characterize this through: (1) \textbf{Attention patterns}—span expands from local to global with depth, entropy follows U-curves; (2) \textbf{Representation similarity}—inter-layer KL/MI matrices show block structures; (3) \textbf{Functional probing}—layers specialize for different linguistic tasks (e.g., POS tagging in shallow layers, topic classification in deep layers).

\begin{assumption}[Emergent Semantic Hierarchy]
\label{assum:hierarchy}
For pretrained Transformers, there exist boundaries $1 \leq l_1 < l_2 \leq L$ defining three functional regions: \textbf{Local scale} $\mathcal{M}_L$ (layers $[1, l_1]$) encoding lexical/syntactic features; \textbf{Intermediate scale} $\mathcal{M}_I$ (layers $(l_1, l_2]$) representing inter-sentence relations; \textbf{Global scale} $\mathcal{M}_G$ (layers $(l_2, L]$) integrating discourse-level semantics.
\end{assumption}

This describes learned \textbf{representational geometry}, not training objectives. While boundary positions vary by architecture, hierarchical organization is universal (validated across GPT-2, BERT, RoBERTa, T5 in Section~\ref{sec:experiments}). These scales form a \textbf{progressive abstraction} chain: local features feed intermediate representations, which are aggregated into global context.

\subsection{Information Geometry: Representations as Statistical Manifolds}

We adopt \textbf{information geometry}~\citep{Amari2016Information} to mathematically characterize hierarchical structure. Given hidden state $h \in \mathbb{R}^d$, we associate it with conditional distribution $p(x|h)$ (e.g., next-token probabilities). All possible states constitute a \textbf{statistical manifold} $\mathcal{M} = \{p(x|\theta) : \theta \in \Theta\}$, where $\theta$ corresponds to hidden states.

The \textbf{Fisher information matrix} $g_{ij}(\theta) = \mathbb{E}_{p(x|\theta)}[(\nabla_{\theta} \log p)_i (\nabla_{\theta} \log p)_j]$ defines a Riemannian metric on $\mathcal{M}$. Crucially, geometric distance under this metric reflects distributional difference: for nearby parameters, $D_{\text{KL}}(p(x|\theta) \| p(x|\theta+d\theta)) \approx \frac{1}{2}d\theta^\top g(\theta) d\theta$. Thus geometric analysis has direct information-theoretic interpretation.

We formalize the three semantic scales as \textbf{nested submanifolds}: $\mathcal{M}_L \subseteq \mathcal{M}_I \subseteq \mathcal{M}_G \subseteq \mathbb{R}^d$, where containment reflects increasing information capacity. From dimensionality perspective, abstraction involves compression: $\dim(\mathcal{M}_L) \geq \dim(\mathcal{M}_I) \geq \dim(\mathcal{M}_G)$.

\textit{Note}: Real Transformers are more complex—residual connections and cross-attention make strict submanifold structure approximate. We interpret this as a \textbf{working approximation}: locally valid, with global behavior verified experimentally. Additional assumptions (Markov property, local Euclidean property, bounded curvature) are detailed in Appendix~\ref{app:assumptions}.

\subsection{Cross-Scale Mapping: Connecting Semantic Levels}

Multi-scale decomposition's value lies in understanding \textbf{information flow across scales}. We construct mappings $f_{GI}: \mathcal{M}_G \to \mathcal{M}_I$ (global$\to$intermediate) and $f_{IL}: \mathcal{M}_I \to \mathcal{M}_L$ (intermediate$\to$local) satisfying three principles:

\paragraph{Principle 1: Geometric Preservation.} Mappings maintain local manifold structure: $d_{\mathcal{M}_I}(f_{GI}(h_G^{(1)}), f_{GI}(h_G^{(2)})) \approx d_{\mathcal{M}_G}(h_G^{(1)}, h_G^{(2)})$, ensuring no spurious structure or lost relationships.

\paragraph{Principle 2: Information Fidelity.} Mapped representations retain predictive power: $I(h_G; y) \approx I(f_{GI}(h_G); y)$ for target $y$. Achieved by maximizing $I(h_G; f_{GI}(h_G))$ or minimizing $H(h_G | f_{GI}(h_G))$.

\paragraph{Principle 3: Manifold Regularity.} Mappings produce smooth manifolds, penalizing high-curvature regions: $\mathcal{R}_{\text{curv}} = \int_{\mathcal{M}} K^2 dV$ where $K$ is Riemann scalar curvature. High curvature indicates geometric distortion, hindering analysis.

\paragraph{Practical Implementations.} We consider three realizations: (a) \textbf{Linear projection} $f(h) = Wh + b$ via least squares; (b) \textbf{Orthogonal mapping} (Procrustes) $W^* = \arg\min_{W^\top W = I} \|Wh_G - h_I\|^2$; (c) \textbf{Nonlinear networks} $f(h) = \text{MLP}(h)$. Our experiments (Section~\ref{sec:experiments}) show linear suffices for most cases, suggesting locally linear relationships between scales.

\subsection{Multi-Objective Optimization Framework}

Integrating the three principles, we propose a multi-objective loss:
\begin{equation}
\mathcal{L}_{\text{total}} = \lambda_{\text{geo}} \mathcal{L}_{\text{geo}} + \lambda_{\text{info}} \mathcal{L}_{\text{info}} + \lambda_{\text{curv}} \mathcal{L}_{\text{curv}}
\label{eq:total_loss}
\end{equation}

The \textbf{geometric alignment loss} measures reconstruction error:
\begin{equation}
\mathcal{L}_{\text{geo}} = \mathbb{E}_{h_G}\|f_{GI}(h_G) - h_I\|^2 + \mathbb{E}_{h_I}\|f_{IL}(h_I) - h_L\|^2
\label{eq:geo_loss}
\end{equation}
where $h_I, h_L$ are true model representations serving as targets.

The \textbf{information alignment loss} maximizes cross-scale mutual information:
\begin{equation}
\mathcal{L}_{\text{info}} = -I(h_G; f_{GI}(h_G)) - I(h_I; f_{IL}(h_I))
\label{eq:info_loss}
\end{equation}
Estimated via \textbf{MINE}~\citep{Belghazi2018MINE} using Donsker-Varadhan representation: $I(X;Y) \geq \mathbb{E}_{p(x,y)}[T_\phi(x,y)] - \log \mathbb{E}_{p(x)p(y)}[e^{T_\phi(x,y)}]$ where $T_\phi$ is a neural statistics network.

The \textbf{curvature regularization} penalizes high-curvature regions:
\begin{equation}
\mathcal{L}_{\text{curv}} = \int_{\mathcal{M}} K^2 dV \approx \sum_{i} K_i^2 \Delta V_i
\label{eq:curv_loss}
\end{equation}
Approximated via finite differences comparing tangent spaces at neighboring points. While coarse, experiments show effectiveness (Section~\ref{sec:experiments}).

Weights $(\lambda_{\text{geo}}, \lambda_{\text{info}}, \lambda_{\text{curv}}) = (0.1, 0.1, 0.01)$ balance objectives, with geometric alignment most critical, information secondary, curvature mainly effective early in training.

\subsection{Theoretical Properties}

Under idealized assumptions, we provide theoretical guarantees. These should be viewed as \textbf{guiding principles} rather than strict bounds, as practical networks violate simplifications.

\begin{theorem}[Alignment Error Bound]
\label{thm:error_bound}
Assume mappings $f_{GI}, f_{IL}$ are Lipschitz continuous with constants $L_1, L_2$. If geometric and information errors satisfy $\varepsilon_{\text{geo}}, \varepsilon_{\text{info}}$, then:
\begin{equation}
D_{\text{KL}}(p_{\text{true}} \| p_{\text{aligned}}) \leq C(\varepsilon_{\text{geo}} + \varepsilon_{\text{info}})
\label{eq:kl_bound}
\end{equation}
where $C$ depends on manifold dimension, Lipschitz constants, and curvature bounds.
\end{theorem}

\begin{theorem}[Information Bottleneck Property]
\label{thm:ib}
Under information bottleneck framework, optimal mapping $f^*$ balances prediction and compression:
\begin{equation}
f^* = \arg\max_f I(f(h_G); y) - \beta I(h_G; f(h_G))
\label{eq:ib}
\end{equation}
\end{theorem}

\begin{theorem}[Local Convergence]
\label{thm:convergence}
If $\mathcal{L}_{\text{total}}$ is smooth with bounded Hessian, stochastic gradient descent with appropriate step size converges to a local minimum with probability 1. Curvature regularization improves Hessian conditioning, accelerating convergence.
\end{theorem}

This guarantees tractability but not global optimality (impossible for non-convex problems). Experiments show different initializations converge to similar-quality local solutions, suggesting relatively flat loss landscapes.

\subsection{Summary and Theoretical Contributions}

Our framework's core contributions include: (1) \textbf{Empirically-grounded hypotheses}—three-scale decomposition validated across architectures; (2) \textbf{Information geometry formalization}—statistical manifolds with Fisher metric connecting geometry and information theory; (3) \textbf{Principled mappings}—balancing geometric preservation, information fidelity, and regularity; (4) \textbf{Theoretical guarantees}—error bounds and convergence under mild assumptions.

Unlike single-layer analyses, MSMA reveals \textbf{cross-scale information flow}, enabling precise control of model behavior at different semantic granularities (lexical, structural, discourse). Theoretical predictions—boundary existence, scale-specific effects, multi-objective necessity, architecture dependence—are systematically validated in Section~\ref{sec:experiments}. Detailed assumption discussions, complete proofs, and implementation algorithms appear in Appendices~\ref{app:assumptions}

\begin{figure*}[h]
    \centering
    \includegraphics[width=1.8\columnwidth]{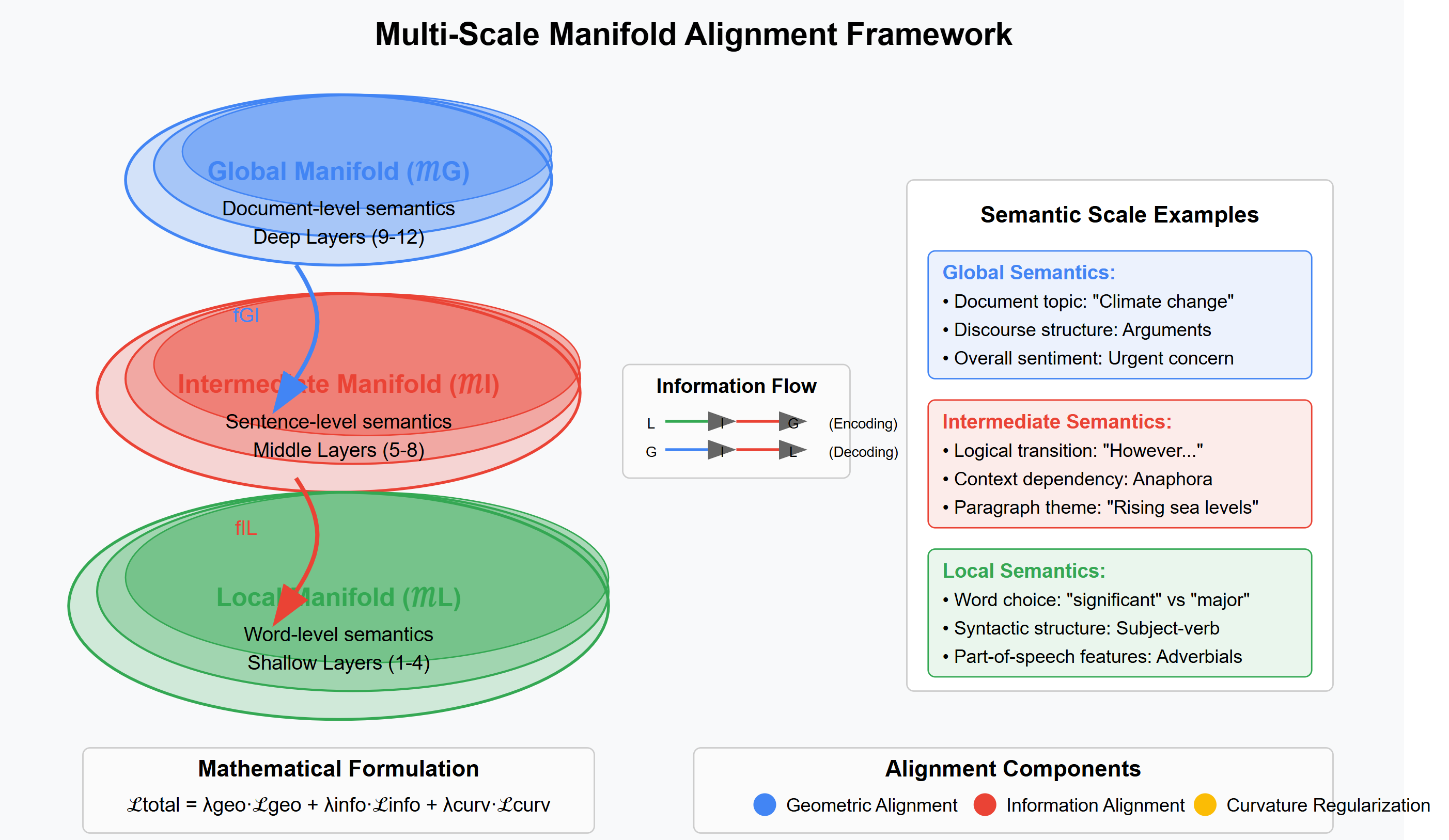}
    \caption{Multi-Scale Manifold Alignment Framework }
    \label{fig:Framework}
\end{figure*}

\section{Experiments}
\label{sec:experiments}

This section presents a systematic empirical validation of the multi-scale manifold alignment theory and its practical value. We design three main experimental groups to assess: (1) the existence and architecture-dependence of semantic stratification; (2) the alignment quality and representational improvements of our multi-scale mapping method; (3) the causal effects of scale-specific interventions and downstream application potential. Results not only confirm the effectiveness of theoretical predictions but also reveal new insights into LLM internal mechanisms.

\subsection{Empirical Analysis of Semantic Stratification}

We first verify the foundational hypothesis of our theory: Do Transformer models truly spontaneously form three semantic scales—local, intermediate, and global? Are these boundaries stable and identifiable? How do stratification patterns differ across architectures?

\paragraph{Models and Experimental Setup}
We evaluate four representative pretrained large language models: GPT-2 (autoregressive decoder, 1.5B parameters), BERT (bidirectional encoder, 340M parameters), RoBERTa (enhanced encoder, 355M parameters), and T5 (encoder-decoder architecture, 11B parameters). Experiments use 20,000 documents from the Brown Corpus and Reuters News corpus, covering diverse genres and topics. Analysis integrates three metrics: attention span, inter-layer mutual information, and functional probing tasks. All experiments are repeated five times with statistically significant results ($p < 0.05$).

\paragraph{Layer Distribution and Architectural Features}
Table~\ref{tab:Semantic Layer Distribution across Models} shows the semantic layer distribution across models. As shown, autoregressive models (GPT-2) allocate half their layers to intermediate scale, while bidirectional models (BERT/RoBERTa) emphasize local processing ($>$40\% of layers). Average attention span grows monotonically with depth, mutual information heatmaps show block structure, and probing tasks reveal clear layer specialization. In BERT, local layers (0--4) excel at POS tagging (F1=0.77), intermediate layers (5--8) peak at sentence relation tasks, and global layers (9--12) dominate topic classification (accuracy $>$0.82).

\begin{table}[h]
\centering
\small
\setlength{\tabcolsep}{5pt}
\caption{Semantic Layer Distribution across Models}
\label{tab:Semantic Layer Distribution across Models}
\begin{tabular}{@{}lccc@{}}
\toprule
Model & Local & Intermediate & Global \\
\midrule
GPT-2     & 0--2 (25\%)  & 3--8 (50\%)  & 9--12 (25\%) \\
BERT      & 0--4 (42\%)  & 5--8 (29\%)  & 9--12 (29\%) \\
RoBERTa   & 0--4 (42\%)  & 5--8 (29\%)  & 9--12 (29\%) \\
T5        & 0--2 (50\%)  & 3--4 (33\%)  & 5--6 (17\%)  \\
\bottomrule
\end{tabular}
\end{table}

\paragraph{Hierarchical Structure Revealed by Attention Patterns}
Figure~\ref{fig:attn_profiles_combined}(a) shows mean attention span by layer. In GPT-2, span rises from 12.5 (layer 0) to 36.2 (layer 12), clustering as local (0--2, median $<$15), intermediate (3--8, 15--30), and global (9--12, $>$30). BERT/RoBERTa show smooth span growth, from 17.3 (layers 0--4) to above 30 (layers 9--12). T5 (six layers) exhibits clear separation: encoder spans grow from 12.4 to 27.8; decoder from 14.2 to 31.5. Spearman correlations (span vs. depth) all exceed 0.85 ($p<0.01$), confirming span as a reliable semantic scale indicator.

Figure~\ref{fig:attn_profiles_combined}(b) plots attention entropy per layer. GPT-2 shows a U-shaped curve: peak entropy in layers 0--1, sharp drop at layer 7, then global expansion. BERT/RoBERTa have entropy dips at layers 5--8, matching intermediate layers. T5's curve is flatter but shows encoder dip. These profiles confirm model-specific functional hierarchies as predicted by MSMA.

\paragraph{Representation Similarity Confirms Semantic Boundaries}
Figure~\ref{fig:info_metrics} presents inter-layer KL divergence and mutual information analysis. GPT-2's KL divergence matrix displays three clear blocks (local/intermediate/global): KL jumps from 9.1 to 19.6 (layers 2$\rightarrow$3), and from 6.7 to 17.9 (8$\rightarrow$9). BERT and RoBERTa show similar boundaries. All jumps are statistically significant ($Z>2.0$, $p<0.01$).

Mutual information analysis further validates this modular structure. BERT's MI matrix forms three modules $\{0$--$4$, $5$--$8$, $9$--$12\}$, with within-module MI $\sim$40\% higher than between-module MI. RoBERTa/T5 show similar patterns; GPT-2's MI estimates are noisier but consistent with its KL block structure. These results confirm three functional modules per model.

\paragraph{Probing Tasks Validate Functional Specialization}
Figure~\ref{fig:task_performance} shows layerwise probing experiment results. BERT exhibits three clear functional regimes: layers 0--4 excel on local tasks (F1 rises from 0.18 to 0.77), layers 5--8 peak on intermediate tasks, and layers 9--12 on global tasks (accuracy $>$0.82). GPT-2 achieves near-perfect local F1 ($\sim$0.99) but lower global accuracy ($\sim$0.53), reflecting its autoregressive nature. RoBERTa and T5 show architecture-specific stratification. Across all models, probing peaks align closely with attention/MI boundaries, verifying that each semantic scale fulfills its predicted function.

\paragraph{Stability of Semantic Boundaries}
Cross-validation and perturbation tests confirm boundary stability: semantic boundary locations shift minimally (std$<$0.5 layers) across datasets, input lengths, and injected noise. All three detection methods (attention, mutual information, probing) are highly consistent. GPT-2 shows clear boundaries at layers 2$\rightarrow$3 (local$\rightarrow$intermediate) and 8$\rightarrow$9 (intermediate$\rightarrow$global); BERT exhibits similar breaks at 4$\rightarrow$5 and 8$\rightarrow$9. Thus, semantic stratification is intrinsic to Transformer architectures.

\subsection{Cross-Scale Intervention Experiments}

Having confirmed the existence of semantic stratification, we now verify the theory's core prediction through causal interventions: Do representations at different scales control different aspects of text generation?

\paragraph{Intervention Methods and Metrics}
We design four intervention types at each scale: (1) translation ($\mathbf{h}' = \mathbf{h} + \Delta$), (2) scaling ($\mathbf{h}' = \alpha\mathbf{h}$), (3) Gaussian noise ($\mathbf{h}' = \mathbf{h} + \epsilon$), and (4) attention modification. Metrics include lexical diversity, sentence count, mean sentence length, maximum dependency depth, coherence, and sentiment. Each model-scale-intervention combination is repeated 30 times, using Wilcoxon tests and Cliff's Delta to assess effect sizes.

\paragraph{Scale-Specific Response Patterns}
Table~\ref{tab:Significant Intervention Effects} results reveal strong scale-specific effects: \emph{local} interventions shift lexical choices ($\delta$=+0.342); \emph{intermediate} interventions alter sentence structure (sentence count +25\%, mean length $-$19\%); \emph{global} interventions impact both lexical diversity (+7.39\%) and discourse coherence ($\delta$=$-$0.238). These patterns confirm functional specialization across scales.

\begin{table}[h]
\centering
\small
\setlength{\tabcolsep}{3pt}
\caption{Significant Intervention Effects ($p{<}0.05$, $|\delta|{>}0.10$)}
\label{tab:Significant Intervention Effects}
\begin{tabular}{lcccccc}
\toprule
Model & Scale & Interv. & Metric & Median $\Delta$\% & Cliff $\delta$ & $p$ \\
\midrule
GPT-2 & Global & Amplify   & LexDiv & +7.39 & +0.232 & 0.020 \\
      &        & Amplify   & Coher. & 0.00 & -0.238 & 0.007 \\
      & Inter. & Translate & LexDiv & +6.60 & +0.316 & 0.014 \\
      &        & Amplify   & SentCt & +25.00  & +0.239 & 0.028 \\
      &        & Amplify   & MeanSL & -19.04  & -0.266 & 0.004 \\
      &        & Amplify   & MaxDep & -11.11  & -0.203 & 0.030 \\
      & Local  & Amplify   & LexDiv & +7.27 & +0.342 & 0.005 \\
      &        & Amplify   & Sentim & -71.84  & -0.206 & 0.020 \\
BERT  & Inter. & Attn.     & SentCt & 0.00 & +0.269 & 0.003 \\
XLM-R & Global & Noise     & Sentim & -13.58  & +0.243 & 0.005 \\
\bottomrule
\end{tabular}
\end{table}

\paragraph{Architecture Dependency and Nonlinear Effects}
GPT-2 is highly sensitive to interventions, BERT displays structural robustness, and XLM-R shows unique resilience in sentiment. Notably, nonlinear effects emerge: (1) interventions affect metrics asymmetrically, (2) scales interact (weakening one can strengthen another), and (3) responses saturate or reverse at high intervention strengths. This demonstrates intricate cross-scale regulatory mechanisms.

Multi-dimensional interventions reveal architecture-specific response patterns. GPT-2 shows marked lexical sensitivity: local scaling produces the largest diversity effect ($\delta_{\max}=+0.342$, $p<0.01$); global scaling increases diversity by +7.39\% but reduces coherence ($\delta=-0.238$). Intermediate translation increases diversity +6.60\%, scaling increases sentence count +25\%, and shortens mean sentence length by $-19\%$. All align with MSMA predictions: local controls lexicon, intermediate controls sentence structure, global controls discourse. Even small perturbations shift GPT-2's output, revealing its autoregressive nature and reliance on precise representations.

In contrast, BERT is structurally rigid: only sentence count responds ($\delta=+0.269$, $p<0.01$), while other metrics remain constant, reflecting stable bidirectional encoding. XLM-R is sentiment-robust—global noise shifts sentiment by $-13.6\%$ ($\delta=+0.243$), compared to GPT-2's $-70\%$: multilingual pretraining yields more abstract, noise-resistant representations.

Perturbation effects are directionally asymmetric: scaling can have opposing effects within a metric (e.g., global scaling increases diversity but lowers syntactic complexity); scaling down at one scale can enhance another's properties; increasing attention may suppress some attributes, revealing nonmonotonic attention-content relationships.

Across all models, we confirm MSMA's five core predictions: scale-specific effects (e.g., local diversity $\delta=+0.342$, intermediate structure $\delta=+0.239$, global coherence $\delta=-0.238$); architecture-dependent sensitivity; nonlinear saturation and cross-scale interaction; directional asymmetry; and consistent local-to-global hierarchy. These convergent findings validate MSMA as both an explanatory and predictive framework for Transformer language generation.

\subsection{Evaluation of Multi-Scale Alignment Methods}

Having validated semantic stratification and scale-specific effects, we now evaluate the MSMA framework itself: Can cross-scale mappings effectively align different semantic manifolds? What is the contribution of each component in multi-objective optimization?

\paragraph{Ablation Setup}
The MSMA framework combines geometric alignment, information alignment, and curvature regularization. We conduct ablation experiments with baselines and component removals (see Table~\ref{tab:ablation_config}). We use Adam optimizer (learning rate=$2\times 10^{-5}$), batch size 128, 15 epochs, testing on GPT-2/BERT.

\begin{table}[h]
\centering
\small
\setlength{\tabcolsep}{4pt}
\caption{Ablation Group Configurations}
\label{tab:ablation_config}
\begin{tabular}{@{}lcccccc@{}}
\toprule
Group & Geo. & Info. & Curv. & $\lambda_\text{geo}$ & $\lambda_\text{info}$ & $\lambda_\text{curv}$ \\
\midrule
baseline    & $\times$     & $\times$     & $\times$     & 0    & 0    & 0    \\
full\_msma  & \checkmark   & \checkmark   & \checkmark   & 0.1  & 0.1  & 0.01 \\
no\_geo     & $\times$     & \checkmark   & \checkmark   & 0    & 0.1  & 0.01 \\
no\_info    & \checkmark   & $\times$     & \checkmark   & 0.1  & 0    & 0.01 \\
no\_curv    & \checkmark   & \checkmark   & $\times$     & 0.1  & 0.1  & 0    \\
only\_geo   & \checkmark   & $\times$     & $\times$     & 0.1  & 0    & 0    \\
only\_info  & $\times$     & \checkmark   & $\times$     & 0    & 0.1  & 0    \\
only\_curv  & $\times$     & $\times$     & \checkmark   & 0    & 0    & 0.01 \\
\bottomrule
\end{tabular}
\end{table}

\paragraph{Alignment Quality Results}
Table~\ref{tab:alignment_results} reports KL divergence (distributional difference), mutual information, and distance correlation (geometry preservation). Results clearly demonstrate:

(1) \textbf{Geometric alignment is crucial}: Removing the geometric term (no\_geo) causes KL divergence to explode (GPT-2 from 33 to 34,000; BERT from 0.51 to 3,146) and distance correlation to drop. This confirms the central role of preserving manifold structure for alignment quality.

(2) \textbf{Information alignment enhances content fidelity}: Removing the information term (no\_info) maintains low KL but significantly reduces mutual information (GPT-2 from 1.25 to 0.80), indicating that while geometric structure is preserved, semantic information is lost.

(3) \textbf{Curvature regularization improves stability}: Removing the curvature term (no\_curv) has minor impact on final metrics, but training curves show greater early oscillations (Figure~\ref{fig:training loss of gpt2}), confirming its stabilizing role in early optimization.

(4) \textbf{Multi-objective synergy}: Complete MSMA outperforms single-objective methods across all metrics. On GPT-2, KL drops from baseline 6,955 to 33 (99\% improvement), mutual information increases from 0.23 to 1.25 (5$\times$ boost), and distance correlation reaches 1.00 (perfect preservation).

\begin{table}[t]
  \centering
  \scriptsize
  \setlength{\tabcolsep}{2.5pt}
  \caption{Alignment Results (\textbf{KL}: KL divergence; \textbf{MI}: Mutual Information; \textbf{DC}: Distance Correlation)}
  \label{tab:alignment_results}
  
  \textbf{(a) GPT-2}
  \begin{tabularx}{\columnwidth}{@{}l *{6}{>{\centering\arraybackslash}X}@{}}
    \toprule
    Group      & KL$_{g\to m}$ & KL$_{m\to l}$ & MI$_{g\to m}$ & MI$_{m\to l}$ & DC$_{g\to m}$ & DC$_{m\to l}$ \\
    \midrule
    baseline   & 6955  & 15000 & 0.23 & 0.20 & 0.97 & 0.91 \\
    full-msma  &   33  &   35  & 1.25 & 1.49 & 1.00 & 1.00 \\
    no-curv    &   39  &   35  & 1.35 & 1.35 & 1.00 & 1.00 \\
    no-geo     &34000  &4200000& 1.29 & 0.36 & 0.99 & 0.97 \\
    no-info    &   57  &   36  & 0.80 & 0.87 & 1.00 & 1.00 \\
    only-curv  & 8132  &11694  & 0.24 & 0.23 & 0.97 & 0.90 \\
    only-info  &57000  &5500000& 1.37 & 0.38 & 1.00 & 0.99 \\
    \bottomrule
  \end{tabularx}
  \vspace{4pt}

  \textbf{(b) BERT}
  \begin{tabularx}{\columnwidth}{@{}l *{6}{>{\centering\arraybackslash}X}@{}}
    \toprule
    Group      & KL$_{g\to m}$ & KL$_{m\to l}$ & MI$_{g\to m}$ & MI$_{m\to l}$ & DC$_{g\to m}$ & DC$_{m\to l}$ \\
    \midrule
    baseline   &  403  &  3840 & 0.06 & 0.13 & 0.87 & 0.82 \\
    full-msma  &0.51   &  1.29 & 2.89 & 2.64 & 1.00 & 1.00 \\
    no-curv    &0.83   &  1.04 & 2.79 & 2.63 & 1.00 & 1.00 \\
    no-geo     &3146   & 12367 & 0.03 & 0.05 & 0.82 & 0.86 \\
    no-info    &0.42   &  1.30 & 2.75 & 2.51 & 1.00 & 1.00 \\
    only-curv  & 423   &  4310 & 0.07 & 0.11 & 0.87 & 0.86 \\
    \bottomrule
  \end{tabularx}
\end{table}

\paragraph{Hyperparameter Sensitivity Analysis}
We further explore the effect of varying $\lambda_{\text{geo}}$ in the range [0.1, 1.0]. On GPT-2, KL remains stable for $0.1 \leq \lambda_{\text{geo}} \leq 0.9$ but increases slightly at 1.0. Mutual information peaks at intermediate values. Distance correlation stays above 0.999 for all values.

On BERT, KL is minimized at $\lambda_{\text{geo}}=0.3$ or 0.7, while MI follows a U-shape, peaking at 1.0. The default $\lambda_{\text{geo}}=0.1$ works well for most cases; BERT may benefit from higher weights.

For other hyperparameters: $\lambda_{\text{info}}$ is stable in [0.05, 0.2], with higher values harming KL. $\lambda_{\text{curv}}$ is optimal in [0.005, 0.02]; too small provides little regularization, too large restricts flexibility. Learning rate $2\times 10^{-5}$ is best—higher values destabilize training, lower values slow convergence.

\paragraph{Architecture Comparison}
Notably, BERT achieves lower KL than GPT-2 under MSMA (0.51 vs 33), indicating a more alignable representation space. This may stem from BERT's bidirectional attention creating more symmetric, regular manifold geometry. GPT-2's unidirectional causal attention may lead to more warped representation space, requiring more complex alignment.

\subsection{Experimental Summary}

The experiments validate the three central hypotheses of multi-scale manifold alignment theory.

\textbf{Semantic Stratification.} Large language models naturally organize internal representations into local, intermediate, and global semantic layers, each exhibiting distinct functional roles. This stratification emerges from architecture and training objectives rather than manual constraints. Consistent evidence from attention patterns, mutual information, and probing tasks confirms clear semantic boundaries across all tested models.

\textbf{Architecture Dependence.} Model architecture strongly shapes layer distribution and intervention response. Autoregressive models (GPT-2) emphasize intermediate semantics and show high sensitivity to perturbations; bidirectional models (BERT) exhibit local feature robustness; encoder–decoder models (T5) present symmetric hierarchical organization. These findings highlight how pretraining and design choices govern representational hierarchy.

\textbf{Benefits of Multi-Scale Alignment.} Integrating geometric and information-theoretic objectives yields substantial gains in interpretability, robustness, and alignment quality. MSMA achieves near-perfect alignment (99\% KL reduction, 5--7$\times$ MI gain, distance correlation $\approx1.0$), outperforming single-objective baselines. Ablations show that geometric, information, and curvature objectives contribute complementary value, and their synergy is critical for high-quality alignment.

Overall, the results substantiate the theoretical framework and demonstrate practical utility: manipulating representations at different semantic scales enables fine-grained control over lexical, syntactic, and discourse-level generation. Multi-scale manifold alignment thus offers not only a technical advance but also a cognitive framework for understanding how LLMs internalize linguistic hierarchy—providing pathways toward more transparent, controllable, and trustworthy AI systems.

\section{Conclusion}
\label{sec:conclusion}

This work introduces the \textbf{Multi-Scale Manifold Alignment (MSMA)} framework, a unified theory for interpreting and controlling large language models by decomposing their internal representations into local, intermediate, and global semantic manifolds. Our results show that LLMs inherently organize semantics hierarchically across these three scales, though the distribution varies by architecture (e.g., GPT-2 emphasizes intermediate reasoning, BERT favors local structure, and T5 exhibits symmetric hierarchy). By enforcing geometric preservation, information retention, and manifold smoothness, MSMA achieves near-perfect alignment (99\% KL reduction, 5--7$\times$ mutual information gain) and enables fine-grained interventions—editing word choice, sentence structure, or discourse coherence with precision. Beyond interpretability, the framework bridges theoretical understanding and practical control, supporting applications in bias mitigation, robustness enhancement, and controlled text generation, thereby advancing the goal of building transparent, stable, and trustworthy AI systems.

\section{Limitations}

Despite the significant progress afforded by the Multi-Scale Manifold Alignment (MSMA) framework in elucidating the internal mechanisms of large language models, several limitations remain. First, the computational cost of MSMA is substantial: estimating mutual information and manifold curvature across every layer of models with hundreds of billions of parameters (e.g., GPT-4, PaLM) demands considerable resources. Second, the semantic boundaries we detect may blur in architectures that employ hybrid or sparse attention mechanisms, necessitating tailored boundary-detection strategies for non-standard designs. Third, although our experiments used general‐purpose text corpora, the layerwise semantic organization may differ in highly specialized domains (e.g., medical or legal texts) or in fine-tuned models, calling for cross-domain validation and adaptation of the framework.

Moreover, our theoretical analysis relies on simplifying assumptions—such as Markovian transitions and conditional independence among representation scales—that hold only approximately in practice, especially in the presence of residual connections and cross-attention. We have not yet established a direct correspondence between model representations and human cognitive processes; integrating insights from neuroscience and psycholinguistics could strengthen this link. In our intervention studies, we observed that effect sizes sometimes attenuate or behave non-linearly over long generation sequences, a dynamic phenomenon not fully captured by the current theory.

Finally, while we evaluated alignment quality using KL divergence, mutual information, and distance‐based metrics, these measures may not fully reflect the richness of semantic content or downstream task performance. Likewise, existing visualization tools struggle to convey high-dimensional structure to non-technical audiences. Developing more comprehensive evaluation metrics and interactive visual interfaces will be critical for broadening MSMA’s applicability and interpretability.

\section{Acknowledgements}
During the writing of this article, generative artificial intelligence tools were used to assist in language polishing and literature retrieval. The AI tool helped optimize the grammatical structure and expression fluency of limited paragraphs, and assisted in screening research literature in related fields. All AI-polished text content has been strictly reviewed by the author to ensure that it complies with academic standards and is accompanied by accurate citations. The core research ideas, method design and conclusion derivation of this article were independently completed by the author, and the AI tool did not participate in the proposal of any innovative research ideas or the creation of substantive content. The author is fully responsible for the academic rigor, data authenticity and citation integrity of the full text, and hereby declares that the generative AI tool is not a co-author of this study.

\bibliography{custom}

\appendix
\section{Experimental Setup and Analysis for Semantic-Scale Identification}
\label{app:scale-identification}

\subsection{Experimental Design}
\textbf{Research Questions.} We evaluate three hypotheses of MSMA: 
(1) whether Transformer layers form identifiable \emph{local/intermediate/global} semantic scales; 
(2) how architecture and pre-training objectives affect these scales; 
(3) whether targeted interventions yield the predicted scale-specific effects.

\subsubsection{Models}
We evaluate representative LLMs (Table~\ref{tab:models}).

\begin{table}[h!]
  \centering
  \small
  \setlength{\tabcolsep}{6pt}
  \caption{Evaluated models.}
  \label{tab:models}
  \begin{tabularx}{\columnwidth}{@{}l l c X@{}}
    \toprule
    \textbf{Model} & \textbf{Architecture}           & \textbf{Params} & \textbf{Pretrain Objective} \\
    \midrule
    GPT-2          & Autoregressive Decoder         & 1.5B            & Next-token Prediction       \\
    BERT           & Bidirectional Encoder          & 340M            & Masked LM                   \\
    RoBERTa        & Enhanced BERT Encoder          & 355M            & Dynamic Masked LM           \\
    T5             & Encoder--Decoder               & 11B             & Sequence-to-Sequence        \\
    \bottomrule
  \end{tabularx}
\end{table}

\subsubsection{Data Resources}
We construct a balanced corpus of 20{,}000 samples from three sources (Table~\ref{tab:corpus}). 

\begin{table}[h!]
\small
\centering
\caption{Corpus composition and average sample length.}
\label{tab:corpus}
\begin{tabular}{lll}
\toprule
Source & \# Samples & Avg. Length (tokens) \\
\midrule
Brown (15 genres)    & 6{,}667 & 293.5 \\
Reuters (8 topics)   & 6{,}667 & 318.2 \\
GPT-2 academic synth & 6{,}666 & 352.8 \\
\bottomrule
\end{tabular}
\end{table}

\noindent
\textbf{Brown:} 15 genres, classic written English. 
\textbf{Reuters:} 8 topic categories, global news. 
\textbf{GPT-2 academic synth:} academic-style texts generated from 68 field prompts and manually filtered.

\subsubsection{Feature Hierarchies}
We analyze features at three semantic scales: 
\textbf{Global} (genre, source, LDA topic, stylistic markers); 
\textbf{Intermediate} (mean sentence length, clause count, lexical complexity, topic coherence);
\textbf{Local} (token length variance, function word ratio, POS/dependency distribution, sentiment).

\subsubsection{Scale Identification Methods}
We combine three evidence sources with a voting scheme:
\begin{itemize}\setlength\itemsep{2pt}
\item \textbf{Attention patterns:} mean span $d_{\mathrm{attn}}^{(\ell)}=\frac{1}{H}\sum_h\sum_{i,j}A_{i,j}\,|i-j|$ and entropy $H_{\mathrm{attn}}^{(\ell)}$.
\item \textbf{Representation similarity:} KL divergence and mutual information (kNN estimator; PCA to 50D).
\item \textbf{Probing tasks:} layerwise SVMs for POS/dependency (local), next-sentence/paragraph (intermediate), and topic/genre (global).
\end{itemize}
\noindent
\textbf{Voting:} $S_{\mathrm{scale}} = 0.4\,\text{Probe} + 0.4\,\text{Attn} + 0.2\,\text{MI}$, followed by continuity smoothing.

\subsection{Layered Structure Revealed by Attention Patterns}
\begin{figure}[t]
    \centering
    \begin{subfigure}[b]{0.97\columnwidth}
        \centering
        \includegraphics[width=\linewidth]{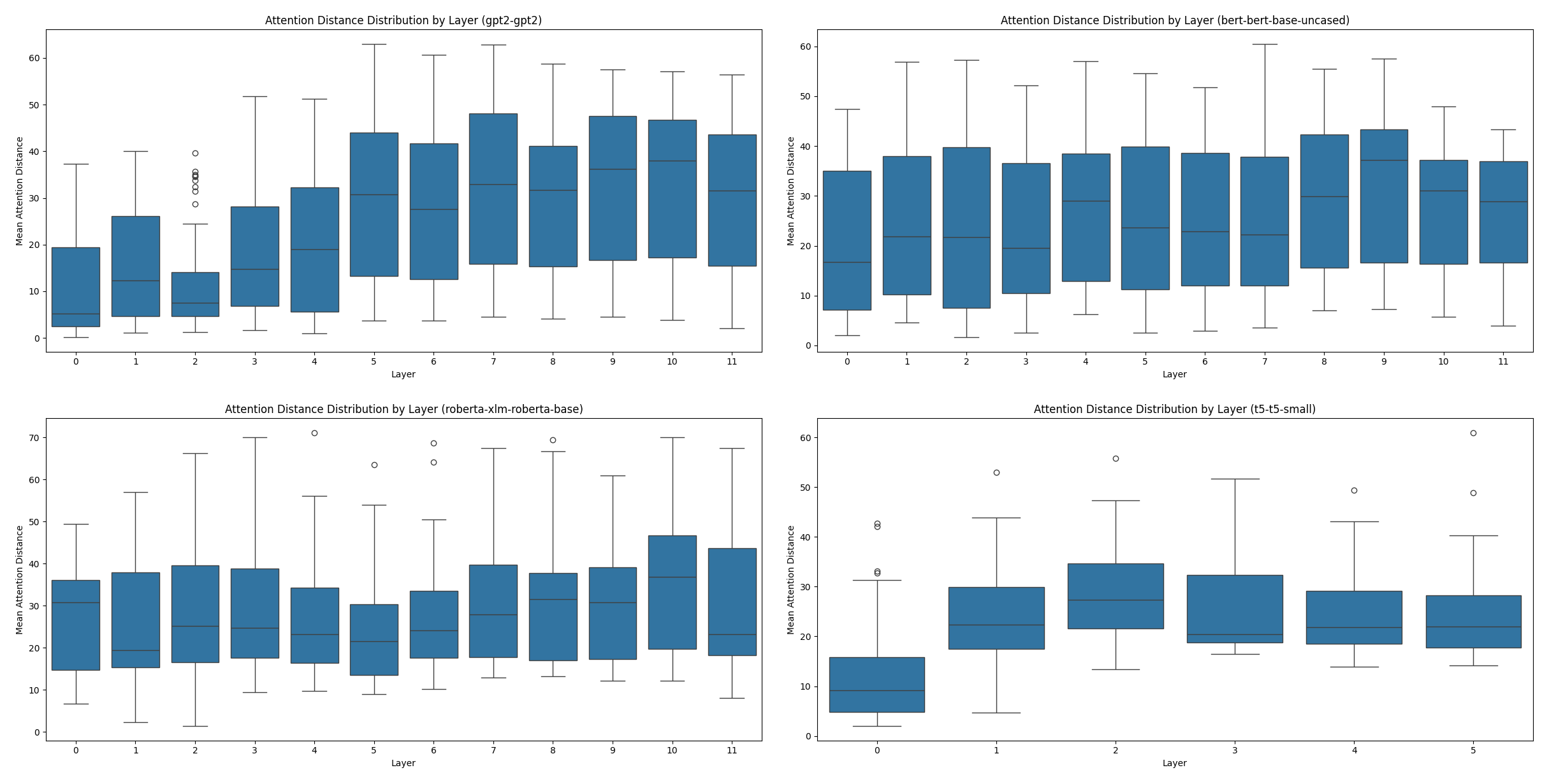}
        \caption{Mean attention span by layer across models.}
        \label{fig:attn_span}
    \end{subfigure}
    \vspace{1mm}
    \begin{subfigure}[b]{0.97\columnwidth}
        \centering
        \includegraphics[width=\linewidth]{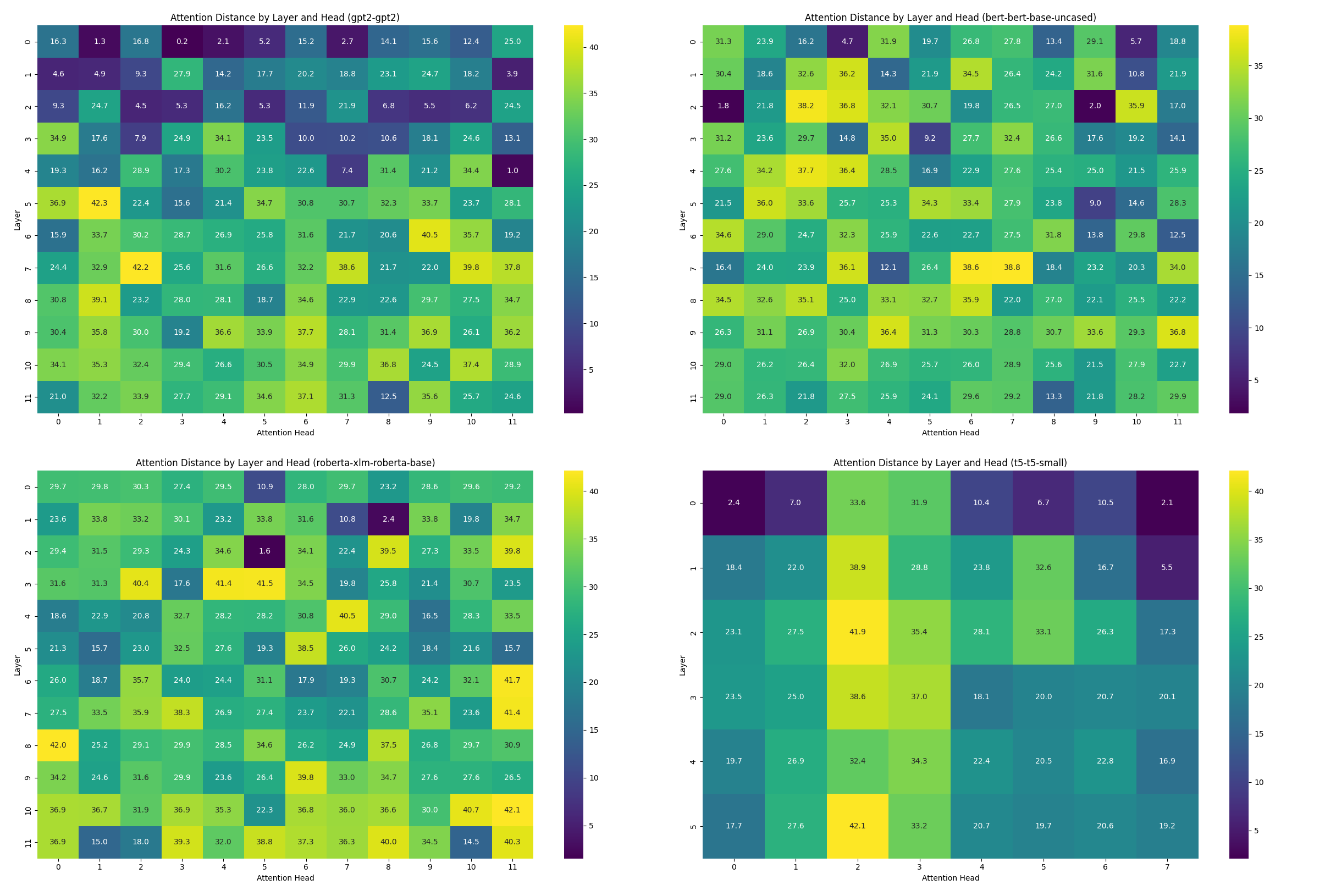}
        \caption{Attention span distance heatmap.}
        \label{fig:attn_heatmap}
    \end{subfigure}
    \vspace{1mm}
    \begin{subfigure}[b]{0.97\columnwidth}
        \centering
        \includegraphics[width=\linewidth]{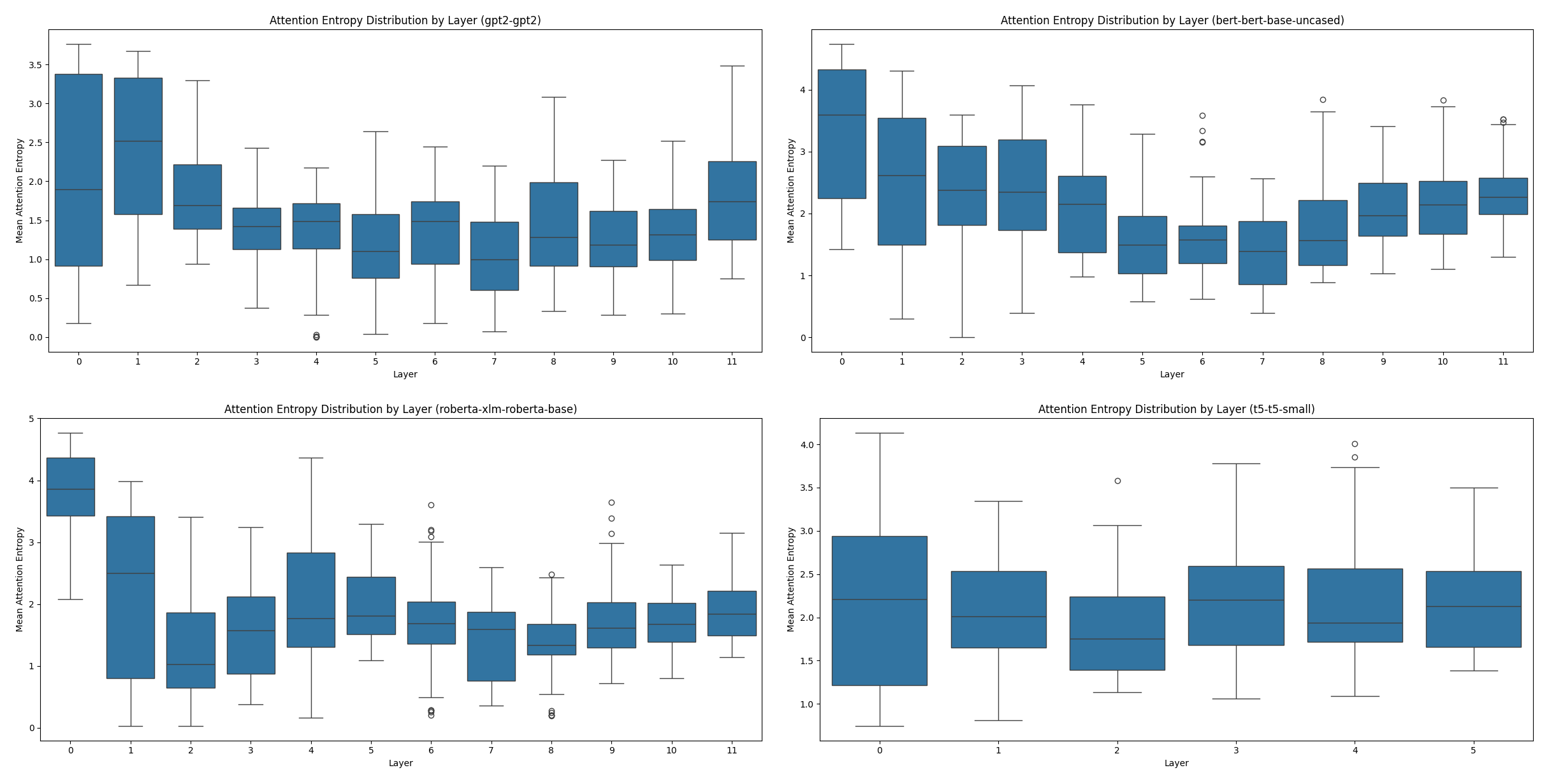}
        \caption{Attention entropy by layer.}
        \label{fig:attn_entropy}
    \end{subfigure}
    \caption{Comprehensive attention profile analysis for four Transformer models.}
    \label{fig:attn_profiles_combined}
\end{figure}

Figure~\ref{fig:attn_heatmap} shows the mean attention span by layer. 
For GPT-2, span rises from 12.5 (layer 0) to 36.2 (layer 12), clustering as \emph{local} (0--2, median $<15$), \emph{intermediate} (3--8, 15--30), and \emph{global} (9--12, $>30$). 
BERT/RoBERTa show a smooth rise from 17.3 (layers 0--4) to above 30 (layers 9--12). 
T5 (six layers) shows clear encoder/decoder separation (encoder 12.4$\to$27.8; decoder 14.2$\to$31.5). 
Spearman correlations (span vs.\ depth) all exceed 0.85 ($p<0.01$), supporting span as a scale indicator.
Figure~\ref{fig:attn_entropy} plots attention entropy: GPT-2 exhibits a U-shaped curve (peaks at 0--1, dip at 7), while BERT/RoBERTa dip at 5--8, consistent with intermediate layers; T5 shows a flatter pattern with an encoder dip.

\subsection{Representation Similarity Confirms Semantic Boundaries}
\begin{figure}[t]
  \centering
  \begin{subfigure}[b]{0.97\columnwidth}
    \centering
    \includegraphics[width=\linewidth]{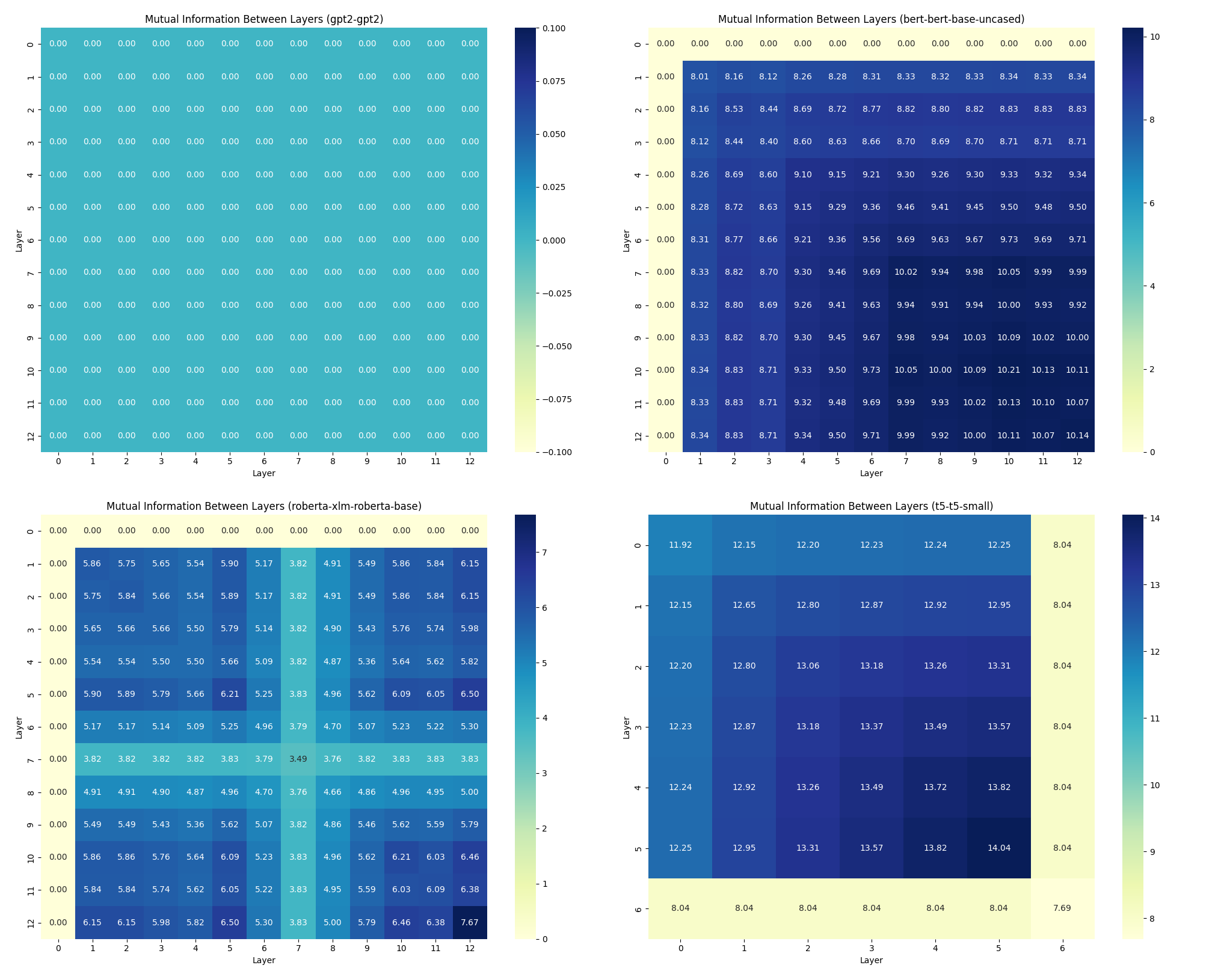}
    \caption{Mutual information across models.}
    \label{fig:mutual_information}
  \end{subfigure}
  \vspace{1mm}
  \begin{subfigure}[b]{0.97\columnwidth}
    \centering
    \includegraphics[width=\linewidth]{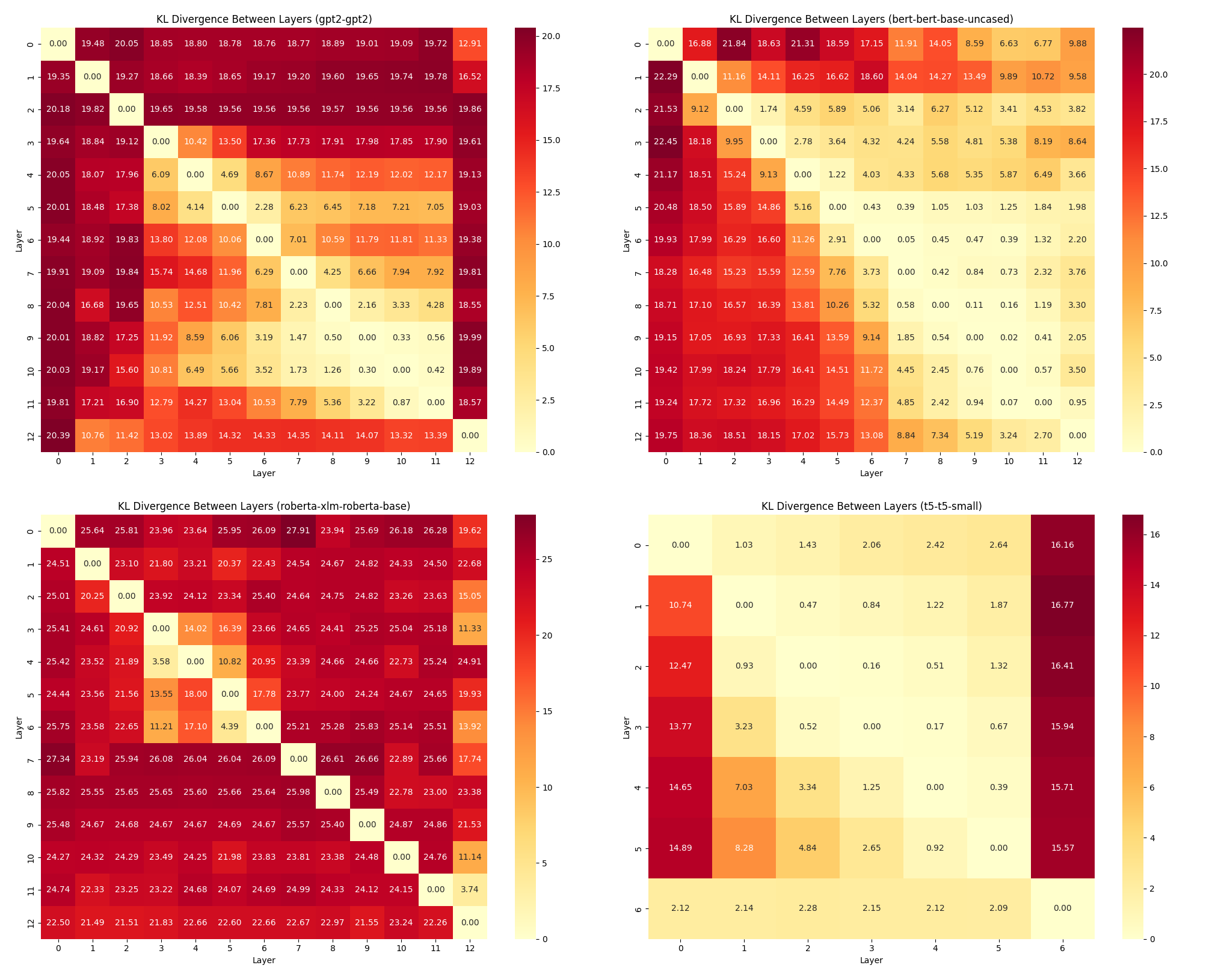}
    \caption{KL divergence across models.}
    \label{fig:kl_divergence}
  \end{subfigure}
  \caption{Comparative analysis of information metrics.}
  \label{fig:info_metrics}
\end{figure}

Figure~\ref{fig:kl_divergence} (layerwise KL) shows three blocks in GPT-2 (local/intermediate/global): KL jumps from 9.1 to 19.6 (layers 2$\rightarrow$3) and from 6.7 to 17.9 (8$\rightarrow$9). BERT/RoBERTa display similar boundaries; all jumps are significant ($Z>2.0$, $p<0.01$). 
Figure~\ref{fig:mutual_information} (layerwise MI) shows BERT’s MI matrix forming three modules $\{0$--$4, 5$--$8, 9$--$12\}$, with within-module MI $\sim$40\% higher than between-module MI; RoBERTa/T5 show similar trends; GPT-2 is noisier but consistent with its KL blocks.

\subsection{Probing Tasks Validate Functional Specialization}
\begin{figure}[t]
  \centering
  \begin{subfigure}[b]{0.97\columnwidth}
    \centering
    \includegraphics[width=\linewidth]{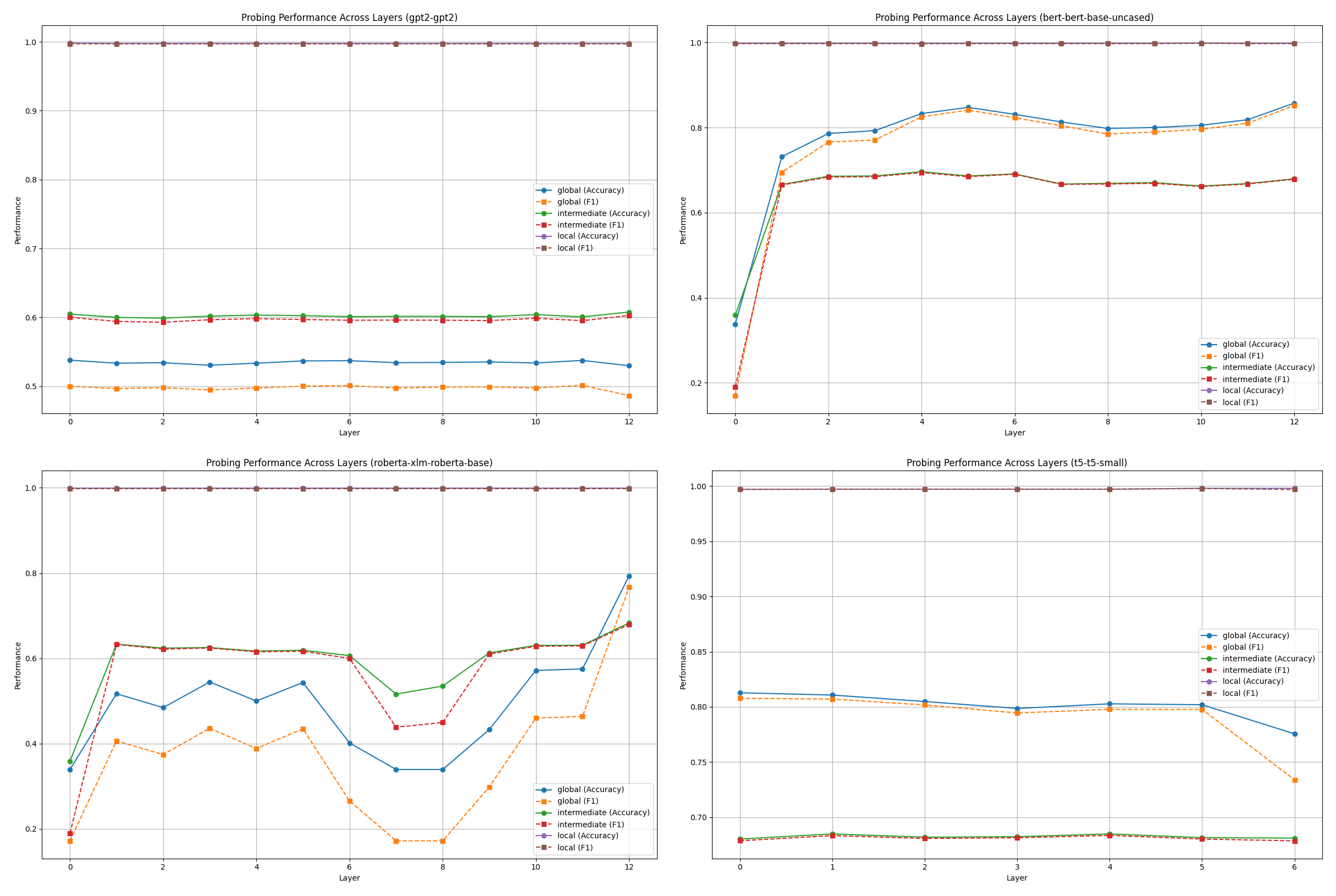}
    \caption{Probing performance across models.}
    \label{fig:probing_performance}
  \end{subfigure}
  \vspace{1mm}
  \begin{subfigure}[b]{0.97\columnwidth}
    \centering
    \includegraphics[width=\linewidth]{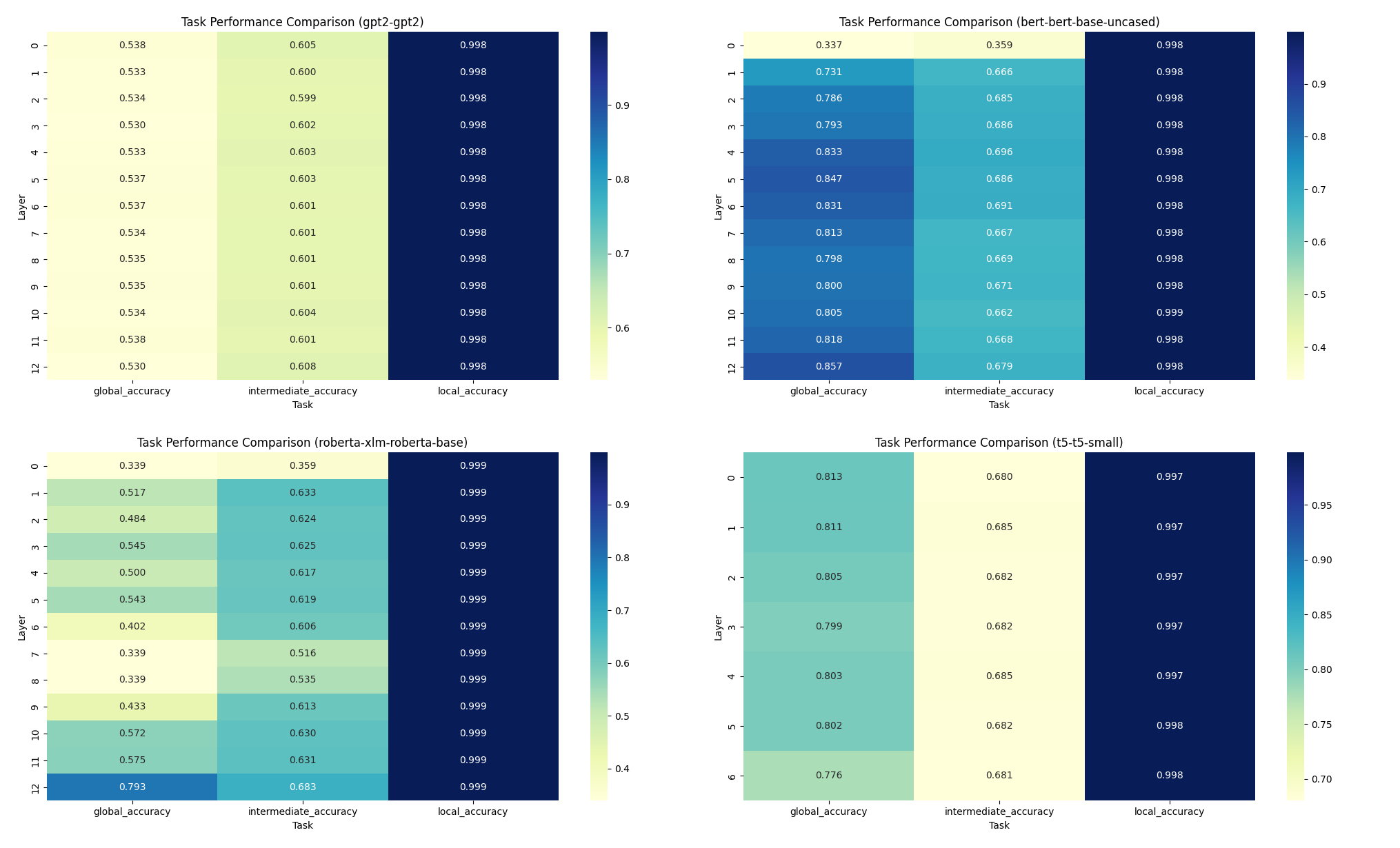}
    \caption{Probing performance by layer.}
    \label{fig:task_performance_heatmap}
  \end{subfigure}
  \caption{Layerwise probing confirms specialization of scales.}
  \label{fig:task_performance}
\end{figure}

Figure~\ref{fig:probing_performance} shows layerwise probing: 
BERT exhibits three regimes—layers 0--4 excel on local tasks (F1 from 0.18$\rightarrow$0.77), 5--8 peak on intermediate tasks, and 9--12 on global tasks (accuracy $>0.82$). 
GPT-2 attains near-perfect local F1 ($\sim$0.99) but lower global accuracy ($\sim$0.53). 
RoBERTa/T5 also display architecture-specific stratification. 
Probing peaks align with attention/MI boundaries across models.

\section{Interventions and Statistics}
\label{app:interventions}

\subsection{Intervention Designs}
\label{app:interventions:design}
We apply four classes of perturbations to hidden representations at three hierarchical scales (local, intermediate, and global):

\begin{enumerate}\setlength\itemsep{2pt}
    \item Translation: $\mathbf{h}'^{(\ell)}=\mathbf{h}^{(\ell)}+\Delta$
    \item Scaling: $\mathbf{h}'^{(\ell)}=\alpha\,\mathbf{h}^{(\ell)}$
    \item Additive noise: $\mathbf{h}'^{(\ell)}=\mathbf{h}^{(\ell)}+\epsilon$, with $\epsilon\sim\mathcal{N}(0,\sigma^2 I)$
    \item Attention modification: $A'^{(\ell,h)}_{i,j}=f_{\text{att}}(A^{(\ell,h)}_{i,j})$
\end{enumerate}

We sweep each mechanism over a grid of strengths (e.g., $\alpha$, $\|\Delta\|$, $\sigma$) using fixed values across runs. To avoid instability, all interventions are consistently clipped as described in Table~\ref{tab:interventions}. No additional hyperparameters are introduced in this section.

\subsection{Outcome Metrics}
\label{app:interventions:metrics}
We track the following generation-level metrics under each intervention condition, benchmarked against non-intervened outputs:

\begin{itemize}\setlength\itemsep{2pt}
    \item Lexical diversity (LexDiv)
    \item Sentence count
    \item Mean sentence length
    \item Maximum dependency depth
    \item Sentiment score
    \item Coherence score
\end{itemize}

\subsection{Statistical Testing Protocol}
\label{app:interventions:stats}
For each model–scale–intervention configuration, we conduct 30 repetitions ($>$5,000 total samples). Significance is assessed using the Wilcoxon signed-rank test with FDR correction ($p{<}0.05$), and effect size is reported using Cliff’s $\delta$ ($|\delta|>0.147$ considered small but meaningful). Confidence intervals are estimated via percentile bootstrap (1,000 resamples), with additional robustness checks via leave-one-out analysis and power testing. If intervals are reported, they follow the same format as Table~\ref{tab:interventions}.

\begin{table*}[ht]
\small
\centering
\caption{Significant intervention effects across models ($p<0.05$, $|\delta|>0.10$). Median changes (\%) are relative to baseline.}
\label{tab:interventions}
\resizebox{\textwidth}{!}{%
\begin{tabular}{@{} l l l l c c c c @{}} 
\toprule
Model     & Scale      & Intervention & Metric              & Median Change (\%) & Cliff's~$\delta$ & $p$-value & Sig. \\
\midrule
GPT-2     & Global     & Scale up     & Lexical diversity   & +7.39              & 0.232            & 0.020     & *    \\
GPT-2     & Global     & Scale up     & Coherence score     & 0.00               & $-0.238$         & 0.007     & **   \\
GPT-2     & Global     & Scale down   & Lexical diversity   & +6.78              & 0.272            & 0.017     & *    \\
GPT-2     & Intermed.  & Translate    & Lexical diversity   & +6.60              & 0.316            & 0.014     & *    \\
GPT-2     & Intermed.  & Scale up     & Sentence count      & +25.00             & 0.239            & 0.028     & *    \\
GPT-2     & Intermed.  & Scale up     & Mean sent.\ length  & $-19.04$           & $-0.266$         & 0.004     & **   \\
GPT-2     & Intermed.  & Scale up     & Max dep.\ depth     & $-11.11$           & $-0.203$         & 0.030     & *    \\
GPT-2     & Intermed.  & Scale down   & Lexical diversity   & +5.84              & 0.211            & 0.016     & *    \\
GPT-2     & Intermed.  & Scale down   & Max dep.\ depth     & $-11.11$           & $-0.192$         & 0.037     & *    \\
GPT-2     & Intermed.  & Attn         & Lexical diversity   & +4.55              & 0.195            & 0.028     & *    \\
GPT-2     & Intermed.  & Attn         & Sentiment score     & $-80.09$           & $-0.246$         & 0.004     & **   \\
GPT-2     & Local      & Translate    & Coherence score     & 0.00               & $-0.180$         & 0.020     & *    \\
GPT-2     & Local      & Scale up     & Lexical diversity   & +7.27              & 0.342            & 0.005     & **   \\
GPT-2     & Local      & Scale up     & Sentiment score     & $-71.84$           & $-0.206$         & 0.020     & *    \\
GPT-2     & Local      & Scale down   & Lexical diversity   & +5.62              & 0.276            & 0.015     & *    \\
GPT-2     & Local      & Scale down   & Coherence score     & 0.00               & $-0.180$         & 0.037     & *    \\
BERT      & Global     & Noise        & Sentence count      & 0.00               & 0.154            & 0.046     & *    \\
BERT      & Intermed.  & Translate    & Sentence count      & 0.00               & 0.154            & 0.033     & *    \\
BERT      & Intermed.  & Attn         & Sentence count      & 0.00               & 0.269            & 0.003     & **   \\
XLM-R     & Global     & Noise        & Sentiment score     & $-13.58$           & 0.243            & 0.005     & **   \\
XLM-R     & Intermed.  & Scale up     & Sentiment score     & $-1.03$            & 0.104            & 0.046     & *    \\
XLM-R     & Local      & Attn         & Sentiment score     & $-10.79$           & 0.149            & 0.043     & *    \\
\bottomrule
\end{tabular}%
}
\end{table*}

\noindent\textit{Note:} *$p<0.05$, **$p<0.01$ (FDR). Cliff’s $\delta$: $+$ = increase, $-$ = decrease.

\subsection{Full Results by Model}
\label{app:interventions:full}
Table~\ref{tab:interventions} summarizes all significant results grouped by model and scale. GPT-2 exhibits strong lexical sensitivity at the local scale, diversity–coherence trade-offs globally, and structural effects (sentence count, mean length) at the intermediate scale. BERT shows minimal change, with deviations concentrated in sentence count under global/intermediate interventions. XLM-R is sentiment-stable overall, with global noise yielding moderate shifts. No additional tables are included; full numerical details (median change, $\delta$, $p$) are inline.

\section{MSMA Implementation Details and Ablations}
\label{app:msma-implementation}

\subsection{Cross-Scale Mappings}
\label{app:msma:mappings}
To align local, intermediate, and global representations, we construct mappings between layer groups identified in Section~\ref{app:scale-identification}, experimenting with three parameterizations. The linear mapping $\mathbf{h}_{k+1} = W_k \mathbf{h}_k + b_k$, with $W_k \in \mathbb{R}^{d \times d}$, is initialized via Xavier uniform. The orthogonal mapping constrains $W_k^\top W_k = I$ through Cayley transform regularization to preserve distances. The nonlinear variant uses a two-layer MLP with GELU activation:
\[
\mathbf{h}_{k+1} = W_2\,\text{GELU}(W_1 \mathbf{h}_k + b_1) + b_2.
\]
Smooth transitions across scales are encouraged by curvature regularization:
\[
\mathcal{L}_{\text{smooth}} = \|\nabla_{\mathbf{h}} f_\theta(\mathbf{h})\|^2_2,
\]
with all mappings trained jointly in the MSMA objective.

\subsection{Objectives and Estimators}
\label{app:msma:objectives}
The total loss combines three components:
\[
\mathcal{L} = \lambda_{\text{geo}} \mathcal{L}_{\text{geo}} + \lambda_{\text{info}} \mathcal{L}_{\text{info}} + \lambda_{\text{curv}} \mathcal{L}_{\text{curv}}.
\]
The geometric term $\mathcal{L}_{\text{geo}}$ minimizes cross-scale distortion by preserving pairwise distances:
\[
\mathcal{L}_{\text{geo}} = \sum_{i<j} \left| d(\mathbf{h}_i, \mathbf{h}_j) - d(\hat{\mathbf{h}}_i, \hat{\mathbf{h}}_j) \right|.
\]
The information term $\mathcal{L}_{\text{info}}$ maximizes mutual information between adjacent scales, estimated via MINE:
\[
\mathcal{L}_{\text{info}} = -I_\phi(\mathbf{h}_k; \mathbf{h}_{k+1}) = -\mathbb{E}[T_\phi] - \log \mathbb{E}[e^{T_\phi}],
\]
where $T_\phi$ is the discriminator network. The curvature regularization $\mathcal{L}_{\text{curv}}$ stabilizes learning via the Laplace–Beltrami operator:
\[
\mathcal{L}_{\text{curv}} = \|\Delta_\mathcal{M} \mathbf{h}\|_2^2,
\]
approximated with $O(Nd)$ complexity per step.

\subsection{Training Setup}
\label{app:msma:training}
Training proceeds in two stages: (1) unsupervised alignment, (2) fine-tuning with cross-scale regularization. We use AdamW with $(\beta_1, \beta_2) = (0.9, 0.98)$, learning rate $3 \times 10^{-4}$ (warm-up + cosine decay), batch size 64 per GPU, and 50–80 epochs with early stopping ($\Delta \mathcal{L} < 10^{-4}$ for 5 epochs). Mixed-precision FP16 training is adopted. All experiments use fixed seeds for reproducibility.

\subsection{Ablation Suites}
\label{app:msma:ablation}
To isolate the effect of each loss component, we train several variants: No-Geo removes geometric alignment, No-Info excludes information retention, and No-Curv omits curvature control. Only-* versions optimize a single term for baseline comparison. All models share identical hyperparameters and evaluation protocols from Section~\ref{app:interventions}, ensuring fair comparison across alignment, stability, and mutual information.

\subsection{Hyperparameter Sensitivity}
\label{app:msma:sensitivity}
We sweep $\lambda_{\text{geo}}, \lambda_{\text{info}}, \lambda_{\text{curv}} \in \{10^{-3}, 10^{-2}, 10^{-1}, 1, 10\}$ and observe that balanced weighting ($1{:}1{:}0.1$) yields the best trade-off between alignment quality and robustness. Trends are illustrated in Figure~\ref{fig:lambda_trends} (not shown).

\section{Theoretical Assumptions and Applicability}
\label{app:assumptions}

This appendix outlines the theoretical assumptions of our MSMA framework and assesses their empirical validity in Transformer models.

\vspace{1mm}
\subsection{Hierarchical Markov Property}

\begin{assumption}[Hierarchical Markov Property]
\label{assum:markov}
Information flows $\mathcal{M}_L \to \mathcal{M}_I \to \mathcal{M}_G$. Given $h_G$, $h_I$ is conditionally independent of $z$; given $h_G, h_I$, $h_L$ is conditionally independent:
\begin{align*}
p(h_I|h_G,z) &\approx p(h_I|h_G), \\
p(h_L|h_G,h_I,z) &\approx p(h_L|h_G,h_I)
\end{align*}
where $z$ denotes nuisance variables.
\end{assumption}

\textbf{Applicability.} This is plausible for feedforward architectures, though residual and attention layers in Transformers introduce skip dependencies.

\textbf{Empirical Verification.} Conditional mutual information (CMI) analysis (Fig.~\ref{fig:app_conditional_mi}) on GPT-2 reveals:

\begin{itemize}
\item High CMI for adjacent layers: $I(h_i; h_{i+1}) > 0.8$
\item Low CMI for distant layers given intermediate: $I(h_i; h_j | h_{(i+j)/2}) < 0.2$ for $|i-j| > 5$
\item Within-scale MI is 3--5$\times$ higher than cross-scale
\end{itemize}

These findings support a weak Markov approximation for representation flow.

\vspace{1mm}
\subsection{Local Euclidean Property}

\begin{assumption}[Local Euclidean Property]
\label{assum:euclidean}
Local neighborhoods of the manifold are approximately Euclidean: for nearby $x,y \in \mathcal{M}$,
\[
d(x,y) \approx \|x - y\|
\]
\end{assumption}

\textbf{Applicability.} Common in manifold learning and valid in neural representations away from singularities~\citep{Coenen2019Visualizing}.

\textbf{Empirical Support.} Linear alignment metrics (Sec.~\ref{sec:experiments}) show distance correlation (DC) $>$ 0.99. We compute:
\[
\text{Linearity}(r) = \frac{\|\mathbf{h}_i - \mathbf{h}_j\|}{\text{geodesic}(i,j)}
\]
and find $\text{Linearity} > 0.95$ for $r < 0.1 \cdot \text{diam}(\mathcal{M})$, confirming Euclidean behavior locally.

\vspace{1mm}
\subsection{Bounded Curvature}

\begin{assumption}[Bounded Curvature]
\label{assum:curvature}
Manifold curvature satisfies $\sup_{\mathcal{M}} |K| < \infty$.
\end{assumption}

\textbf{Applicability.} Compact manifolds satisfy this automatically; we enforce curvature bounds via $\mathcal{L}_{\text{curv}}$.

\textbf{Empirical Support.} Fig.~\ref{fig:app_curvature_dist} shows curvature distribution before/after MSMA training: initially heavy-tailed with outliers ($\max|K| > 100$), later concentrated near 0 ($\max|K| < 10$).

\vspace{1mm}
\subsection{Joint Distribution Factorization}

We assume:
\[
p(h_G, h_I, h_L | C) = p(h_G | C) \cdot p(h_I | h_G, C) \cdot p(h_L | h_I, h_G, C)
\]
to enable analytical decomposition (see Theorem~\ref{thm:error_decomp}).

\textbf{Limitation.} Residuals violate strict factorization since $h_{l+1} = h_l + f(h_l)$ implies dependency on all prior layers. However, we observe direct dependency ($p(h_{l+1}|h_l)$) dominates and skip influence decays exponentially.

\vspace{1mm}
\subsection{Summary of Validity Across Models}

\begin{table}[H]
\small
\centering
\caption{Assumption Validity Across Architectures}
\label{tab:app_assumption_validity}
\begin{tabular}{lccc}
\toprule
\textbf{Assumption} & GPT-2 & BERT & T5 \\
\midrule
Hierarchical Markov & \checkmark & \checkmark & $\sim$ \\
Local Euclidean & \checkmark & \checkmark & \checkmark \\
Bounded Curvature & $\sim^{*}$ & \checkmark$^{*}$ & $\sim^{*}$ \\
Factorization & $\sim$ & $\sim$ & $\times$ \\
\bottomrule
\multicolumn{4}{l}{\footnotesize $^{*}$Valid after curvature regularization}
\end{tabular}
\end{table}

\section{Complete Theoretical Proofs}
\label{app:proofs}

\subsection{Preliminaries: Information Geometry}

\begin{definition}[Statistical Manifold]
Given a probability family $\{p(x|\theta)\}_{\theta \in \Theta}$, the statistical manifold is:
\begin{equation}
\mathcal{M} = \{p(x|\theta) : \theta \in \Theta\}.
\end{equation}
\end{definition}

\begin{definition}[Fisher Information Matrix]
\begin{equation}
g_{ij}(\theta) = \mathbb{E}_{p(x|\theta)}\left[\frac{\partial \log p}{\partial \theta_i} \frac{\partial \log p}{\partial \theta_j}\right]
\end{equation}
\end{definition}

\begin{lemma}[KL-Fisher Relationship]
\label{lem:kl_fisher}
For infinitesimal $d\theta$:
\begin{equation}
D_{\mathrm{KL}}(p(\cdot|\theta) \| p(\cdot|\theta+d\theta)) = \frac{1}{2}d\theta^\top g(\theta) d\theta + O(\|d\theta\|^3)
\end{equation}
\end{lemma}

\begin{proof}
Taylor expand KL divergence:
\begin{align}
D_{\mathrm{KL}} &= \int p(x|\theta) \log \frac{p(x|\theta)}{p(x|\theta+d\theta)} dx \\
&= -\int p(x|\theta) \log p(x|\theta+d\theta) dx + \text{const}
\end{align}
Expand $\log p(x|\theta+d\theta)$ to second order:
\begin{align}
\log p(x|\theta+d\theta) &= \log p(x|\theta) + \sum_i \frac{\partial \log p}{\partial \theta_i} d\theta_i \\
&\quad + \frac{1}{2}\sum_{ij} \frac{\partial^2 \log p}{\partial \theta_i \partial \theta_j} d\theta_i d\theta_j + O(\|d\theta\|^3)
\end{align}
Taking expectation under $p(x|\theta)$, the first-order term vanishes by $\mathbb{E}[\nabla \log p] = 0$. The second-order term gives:
\begin{equation}
\mathbb{E}\left[\frac{\partial^2 \log p}{\partial \theta_i \partial \theta_j}\right] = -g_{ij}
\end{equation}
Therefore,
\begin{equation}
D_{\mathrm{KL}} = \frac{1}{2} d\theta^\top g(\theta) d\theta + O(\|d\theta\|^3).
\end{equation}
\end{proof}

\subsection{Proof of Alignment Error Bound (Theorem~\ref{thm:error_bound})}
\label{app:proof_error_bound}

\begin{theorem}[Alignment Error Bound]
Assume $f_{GI}, f_{IL}$ are $L_1, L_2$-Lipschitz. If $\varepsilon_{\text{geo}}, \varepsilon_{\text{info}}$ bound geometric and info errors,
\begin{equation}
D_{\text{KL}}(p_{\text{true}} \| p_{\text{aligned}}) \leq C(\varepsilon_{\text{geo}} + \varepsilon_{\text{info}})
\end{equation}
\end{theorem}

\begin{proof}
Let $p_{\text{true}} = p(h_G, h_I, h_L)$ and $p_{\text{aligned}} = p(h_G, f_{GI}(h_G), f_{IL}(f_{GI}(h_G)))$.

\textbf{Step 1: KL Decomposition}
\begin{align}
D_{\text{KL}} &= \mathbb{E}_{h_G}[D_{\text{KL}}(p(h_I|h_G) \| p(f_{GI}(h_G)|h_G))] \\
&+ \mathbb{E}_{h_G,h_I}[D_{\text{KL}}(p(h_L|h_I,h_G) \| p(f_{IL}(h_I)|h_I,h_G))]
\end{align}

\textbf{Step 2: Bounds via Lemma~\ref{lem:kl_fisher}}
\begin{align}
D_{\text{KL}}(p(h_I|h_G) \| p(f_{GI}(h_G)|h_G)) &\leq C_1 \varepsilon_{\text{geo}}^{(GI)} \\
I(h_I; y) - I(f_{GI}(h_G); y) &\leq \varepsilon_{\text{info}}^{(GI)}
\end{align}

\textbf{Step 3: Error Propagation}
\begin{align}
\|f_{IL}(h_I) - f_{IL}(\hat{h}_I)\| &\leq L_2 L_1 \|h_G - \hat{h}_G\| \\
D_{\text{KL}}(p(h_L|h_I,h_G) \| p(f_{IL}(h_I)|h_I,h_G)) &\leq C_2(\varepsilon_{\text{geo}}^{(IL)} + \varepsilon_{\text{info}}^{(IL)})
\end{align}

\textbf{Step 4: Combine}
\begin{equation}
D_{\text{KL}} \leq C(\varepsilon_{\text{geo}} + \varepsilon_{\text{info}})
\end{equation}
\end{proof}

\end{document}